\documentclass[letterpaper, 10 pt, conference]{ieeeconf}  

\IEEEoverridecommandlockouts                              

\usepackage{graphics} 
\usepackage{graphicx}
\usepackage[noend]{algpseudocode}
\usepackage{amsmath}
\usepackage{subcaption}
\usepackage{textcomp}
\usepackage{siunitx}
\usepackage{tikz}
\usepackage[linesnumbered,ruled]{algorithm2e}
\usetikzlibrary{shapes,arrows}
\usepackage{multirow}
\usepackage{romannum}
\makeatletter
\newcommand{\mypm}{\mathbin{\mathpalette\@mypm\relax}}
\newcommand{\@mypm}[2]{\ooalign{%
  \raisebox{.1\height}{$#1+$}\cr
  \smash{\raisebox{-.6\height}{$#1-$}}\cr}}
\makeatother

\def\inv{\vspace*{-0.15cm}}

\title{\LARGE \bf Clumped Nuclei Segmentation with Adjacent Point Match and Local Shape-Based Intensity Analysis in Fluorescence Microscopy Images}

\author{Xiaoyuan Guo, Hanyi Yu, Blair Rossetti, George Teodoro, Daniel Brat and Jun Kong
\thanks{Xiaoyuan Guo, Hanyi Yu, Blair Rossetti and Jun Kong are with the Emory University, Dept. of Biomedical Informatics,  Atlanta, GA 30322 ({\tt\footnotesize \{xiaoyuan.guo, hanyi.yu, blair.rossetti, jun.kong\}@emory.edu}); George Teodoro is with the University of Bras\'{\i}lia, Dept. of Computer Science,  Bras\'{\i}lia, DF, Brazil ({\tt\footnotesize glmteodoro@gmail.com}); Daniel Brat is with the Northwestern University, Dept. of Pathology, Chicago, IL 60611 ({\tt\footnotesize daniel.brat@northwestern.edu});  Funded by NIH K25CA181503, and CNPq.}
 }

\begin{document}

\maketitle
\thispagestyle{empty}
\pagestyle{empty}

\begin{abstract}
Highly clumped nuclei captured in fluorescence microscopy images are commonly observed in a wide spectrum of tissue-related biomedical investigations. To ensure the quality of downstream biomedical analyses, it is essential to accurately segment clustered nuclei. However, this presents a technical challenge as fluorescence intensity alone is often insufficient for recovering the true nuclei boundaries.
In this paper, we propose an segmentation algorithm that identifies point pair connection candidates and evaluates adjacent point connections with a formulated ellipse fitting quality indicator. After connection relationships are determined, we recover the resulting dividing paths by following points with specific eigenvalues from the image Hessian in a constrained searching space. We validate our algorithm with 560 image patches from two classes of tumor regions of seven brain tumor patients. Both qualitative and quantitative experimental results suggest that our algorithm is promising for dividing overlapped nuclei in fluorescence microscopy images widely used in various biomedical research. 
\end{abstract}


\section{INTRODUCTION}~\label{sec:intro}
Fluorescence microscopy images are commonly used in clinical and biomedical research to help visualize cellular components, such as membranes and nuclei. Analysis of fluorescence microscopy images often requires an accurate identification of individual cell nuclei. However, clumped nuclei due to a high cell density often make it challenging to achieve an accurate nuclei segmentation. Although a large number of clumped nuclei segmentation methods have been proposed for fluorescence microscopy image analysis, they are subject to salient defects~\cite{Multifractal2017,Delaunay2009}. For example, shape-based methods are sensitive to clumped nuclei shape variance which often leads to under-segmentation~\cite{Delaunay2009,bottleneck2011,Liao2016,Fouad2016}. Similarly, under-segmentation occurs in concavity-based methods that heavily rely on the clumped nuclei contour concavity~\cite{Fouad2016,curvatureweighting2012,Zhang2013}. By contrast, existing watershed-based methods usually yield over-segmentation due to the variance of image density~\cite{Atta-Fosu,watershed-based2017}. Although ~\cite{Atta-Fosu} applied a 3D analysis to segment the clumped nuclei to enhance the segmentation accuracy, the complexity of the algorithm grows.

We propose a new algorithm that fully exploits the boundary shape of nuclei clumps to identify connecting point pairs by local high curvature voting and point pair screening. This is followed by a novel method that connects point pairs with dividing curves derived from local shape-based intensity analysis. Our method shows significant improvements over an existing marker-controlled watershed method~\cite{watershed-based2017} when tested on a set of 560 fluorescence microscopy image patches.The paper is organized as follows. We present our algorithm in Section~\ref{sec:method} and show the experimental results in Section~\ref{sec:result}.  In Section~\ref{sec:conclusion}, we conclude our paper.

\section{METHOD}~\label{sec:method}
\begin{figure}
\centering
  \includegraphics[width=\linewidth,height=5.5cm]{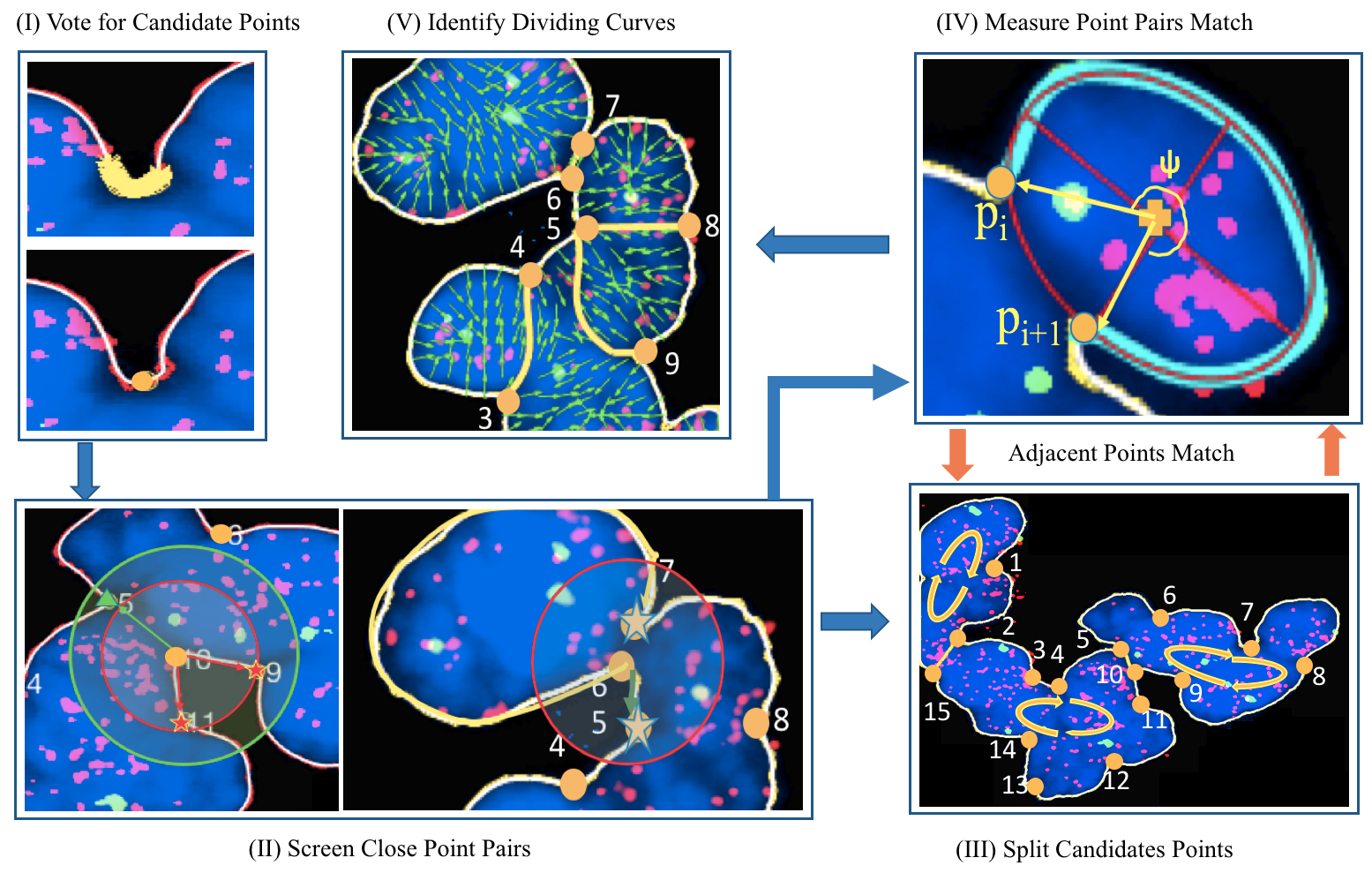}
  \caption{The overall method flowchart for our proposed clumped nuclei segmentation is demonstrated. Nuclei are shown in blue with additional nuclear components shown as red and green interior regions.}
  \label{fig:procedure} \inv\inv
\end{figure}

Our algorithm utilize the contour shape of clustered nuclei to generate candidate point pairs and local shape based intensity analysis to separate overlapped nuclei with dividing curves. The work-flow consists of four main steps: (1) Voting for high curvature candidate points; (2) Screening close point pairs by a curvature-based distance metric for adjacent point pairs and a comprehensive metric taking into account curvature, distance, and matched normal vector angle for non-adjacent point pairs; (3) Assessing adjacent point pair connections via an ellipse fitting quality indicator; and (4) Identifying dividing curves by local shape based intensity analysis (see Fig.~\ref{fig:procedure}).

\subsection{High Curvature Candidate Point Voting}~\label{subsec:curvature}
The blue-fluorescent DAPI nucleic acid stain preferentially binds to DNA within nuclei. Thus, we focus on the blue image channel for overall nuclear contour detection. We obtain the boundaries of such clustered nuclei using a global adaptive threshold. We propose a new method to find high curvature points on clump contours based on neighboring high curvature point votes. Curvature value of each point on the boundary is computed by  $\kappa=\frac{{x}'{y}''-{y}'{x}''}{({x}'^{2}+{y}'^{2})^{3/2}} $, where $x$ and $y$ are coordinates of points on the denoised contour convolved with a smoothing Gaussian filter~\cite{Delaunay2009}. Different from high curvature point detection methods that select points with curvature values higher than a specified threshold~\cite{Delaunay2009}, our method can avoid redundant candidate point detection with local high curvature point votes. As the following segmentation analysis depends on these initial candidate points, limiting the number of redundant candidate points can substantially reduce the computational cost and improve segmentation accuracy. In our data, it is quite common to have a long set of high curvature points along a concave contour segment defined by the curvature sign, as demonstrated in Fig.~\ref{fig:procedure}(\Romannum{1}). For those high curvature points in spatial proximity, it is unreasonable to mark all these points as candidates. Thus, we propose to vote for the optimal candidate point representation for a group of adjacent high curvature points by:
\begin{equation}
s^{\ast}=\frac{\int_{\substack{\partial C}} \kappa(s)s ds}{\int_{\substack{\partial C}} \kappa(s)ds}
\end{equation}
\noindent where $s\in [0, 1]$ represents the normalized boundary arc length on a cell concave contour segment $\partial C $; $\kappa(s)$ is the curvature value at $s$; $s^{\ast}$ is the normalized arc length of the optimal point representation for a segment of adjacent concavity points engaged in local voting. The bottom image in Fig.~\ref{fig:procedure}(\Romannum{1}) demonstrates the final high curvature point representation (in yellow circle) voted by all local high curvature points detected in the top image in Fig.~\ref{fig:procedure}(\Romannum{1}). With this procedure, we detect all high curvature point representations on each contour. 

\subsection{Close Point Pair Screening }~\label{subsection:candidate}
Let us denote $P=\{p_{i},| i=1,2, \ldots, N\}$ as a set of  high curvature candidate points detected from Subsection~\ref{subsec:curvature}.  As demonstrated in Fig.~\ref{fig:procedure}(\Romannum{2}), we next identify all pairs of points $(p_{i},p_{j})$ in spatial proximity by two circular searching regions of radius $r_1$ and $r_2$ ($r_1 < r_2$). Specifically, $r_1$ is used to detect the pairs of both adjacent and non-adjacent points, whereas the ring area defined by $r_2 - r_1$ is for detecting pairs of non-adjacent points in a neighboring area. In our analysis, they are set to 45 pixels and 70 pixels, respectively. In this way, we can alleviate under-segmentation suffered by numerous shape-based methods. The spatial proximity in our study is measured by the Euclidean distance. In addition, unlike the ``bottleneck'' points defined as close points presenting opposite gradient directions in previous research~\cite{bottleneck2011, Liao2016}, we relax this constraint and classify close points as either adjacent or non-adjacent point pairs. Plotted in the left panel in Fig.~\ref{fig:procedure}(\Romannum{2}), the search regions for close point pairs are represented by the red circle with radius $r_1$ and the green ring with inner radius $r_1$ and outer radius $r_2$, respectively. Here we use the searching region in the red circle of radius $r_1$ for detecting pairs of both adjacent and non-adjacent points in spatial proximity, whereas we only use the green ring for detecting pairs of non-adjacent points in a neighboring area. In the left plot of Fig.~\ref{fig:procedure}(\Romannum{2}), points 9 and 11 in the red circle region are close adjacent points of point 10, while point 5 in the green area is also close but not adjacent to point 10. Although these two types of close point pairs are processed in different ways in our proposed method, they both are important for an accurate segmentation.

(1) {\it Analysis for Pairs of Adjacent Points in Proximity}: Let us denote $C^{+} = \{(p_{i},q_{i}) | i=1,2,\ldots,N_1\} $ as the set of neighboring adjacent point pairs given the searching circular area of radius $r_1$. For each such pair demonstrated in the right panel of Fig.~\ref{fig:procedure}(\Romannum{2}), we evaluate the associated Walking Energy to determine if we merge them (i.e. discard the undesired point and keep the point with the higher curvature value). Walking Energy is defined as $E_{p, q}=\int_{\substack{p(s)}}^{q(s)} \kappa(s) ds$, where $\kappa(s)$ is the curvature of the corresponding curve segment $s\in [p(s), q(s)]$. Walking Energy represents the amount of "effort" required for a walk from a point $p(s)$ to its close adjacent point $q(s)$. In our formulation, we assume it consumes more energy to walk along a convex than non-convex contour. Demonstrated in the right panel of Fig.~\ref{fig:procedure}(\Romannum{2}), point 6 has two close point neighbors, i.e. point 5 and 7. The true walking route from point 6 to 7 is illustrated in a red curve, while that from point 6 and 5 is shown in green. Those pairs of neighboring adjacent points requiring low Walking Energy are merged. For each such pair of points under investigation, the point with higher curvature value is retained after points get merged. Adjacent points associated with a high Walking Energy would be retained, as shown in the right panel of Fig.~\ref{fig:procedure}(\Romannum{2}) where the Walking Energy for points 6 and 7 is sufficiently high to keep them separated. For any pair of such adjacent points presenting a high Walking Energy, it is highly likely that such points should be connected to separate overlapped nuclei. Therefore, we further evaluate the boundary segmentation quality in such cases by an ellipse fitting method discussed in Subsection~\ref{subsec:ellipse}.

(2) {\it Analysis for Pairs of Non-Adjacent Points in Proximity}: The resulting set of non-adjacent point pairs in close proximity can be represented as  $C_1^{-} = \{(p_{i},q_{i}) | i=1,2,\ldots,N_2\}$ when we use the circular searching area of radius $r_1$. To alleviate under-segmentation, we next expand the searching space to a ring searching region with radius between $r_1$ and $r_2$. To assess if such pairs of non-adjacent points $(p_i, q_i)$ result from intersection of overlapped nuclei, we formulate and use the following measure to evaluate such point pairs:
\begin{equation}
V(p_i, q_i) = \frac{\alpha \theta(p_i, q_i) }{ D(p_i, q_i)+\beta (\kappa(p_i)+ \kappa(q_i)) } 
\end{equation}
\noindent where $ \theta$ is the angle between two normal vectors of the nuclear contour at $p_i$ and $q_i$; $D$ is the Euclidean distance between $(p_i, q_i)$; $\alpha$ and $\beta$ are weights set as 100 and 0.34, respectively. If $V(p_i, q_i)$ is larger than a specified threshold (e.g. 200 in our analysis), and $(p_i, q_i)$ are non-adjacent points in a ring searching space between the radius of $r_1$ and $r_2$, such point pair is added to $C_2^{-} = \{(p_{i},q_{i}) | i=1,2,\ldots,N_3\}$. 
After this evaluation analysis, the finalized set of non-adjacent point pairs in close proximity is $C^{-} = C_1^{-} \bigcup C_2^{-}$. For each pair in $C^{-}$, we next link the associated points to partition the whole contour into multiple subsets. Demonstrated in Fig.~\ref{fig:procedure}(\Romannum{3}), a whole nuclei cluster can be partitioned into multiple subsets by connecting points $(2, 15)$, and $(5, 10)$. As each candidate subset can produce new possible adjacent point pairs, the method for pairs of adjacent points in proximity discussed above is applied to each candidate  subset. 

\subsection{Point Pair Segregation Assessment by Ellipse Fitting}~\label{subsec:ellipse}
We proceed with evaluating the segregation quality $Q_{i, i+1}$ of each adjacent point pair $(p_{i}, p_{i+1})$ by ellipse fitting. For each sub-contour segment under evaluation, an optimal ellipse, shown in red in Fig.~\ref{fig:procedure}(\Romannum{4}), is fit to a candidate nucleus region with a closed sub-contour consisting of a contour arc and a dividing line  between $(p_{i}, p_{i+1})$. An ellipse fitting quality measurement is defined as:
\begin{equation}
Q_{i,i+1}=\frac{\mu S^{+}_{i, i+1}+\nu \psi_{i, i+1}}{ (\Delta x_{i, i+1}+\Delta y_{i, i+1})+\gamma_{1} \Delta L_{i, i+1}+\gamma_{2} \eta_{i, i+1} }
\end{equation}
\noindent where $ S^{+}_{i, i+1}$ is the ratio of overlapped area between the candidate nucleus and fitting ellipse to their union; $\psi_{i, i+1}$ is the fitting angle formed by the line segment connecting the ellipse center and $p_i$ and that connecting the ellipse center and $p_{i+1}$; $\Delta x_{i, i+1}$ and $\Delta y_{i, i+1}$ are the centroid coordinate difference between the fitting ellipse and the evaluated nucleus region; $\Delta L_{i, i+1}$ is the perimeter difference between the fitting ellipse and the evaluated nucleus contour; $\eta_{i,i+1}$ is the ratio of major to minor axis length derived from the fitting ellipse. $\mu, \nu,\gamma_{1}$ and $\gamma_{2}$ are weights set to 10.70, 10.70, 0.67 and 3.40 in our experiment. 

With the formulated ellipse fitting quality measurement, we connect those neighboring adjacent point pairs when the associated $Q$ is larger than a threshold (e.g. 0.7  in  our  analysis). This ellipse fitting evaluation process is carried out in an iterative manner. Each time, we connect the pair with the largest $Q$. In the post-processing step, we  prune intersected connections and dividing lines that form a sharp angle to avoid over-segmentation. Finally, we obtain pairs of points to be connected in a set $C^{\ast} = \{(p_{i}^\ast,q_{i}^\ast) | i=1,2,\ldots,N_{C}\} $.

\subsection{Identification of Dividing Curves}
Given identified point pairs for connection, a vast majority of nuclei segmentation methods connect each such pair with a straight separating line. However, it is our observation that a better division is given by a curve following the image gradient information, as shown in Fig.~\ref{fig:procedure}(\Romannum{5}). Therefore, we next aim to recover a dividing curve for each such pair with local shape-based intensity analysis. Denoting $I(x,y)$ as the fluorescence microscopy image, we can represent it by Taylor series expansion as:
\begin{align}
& I(x,y)\approx  I(x_{0},y_{0})+\bigtriangledown I(x_{0},y_{0})\begin{bmatrix}
 x-x_0\\ y-y_0 \end{bmatrix} \\ \nonumber
&+\frac{1}{2}\begin{bmatrix}
 x-x_0 &  y-y_0 \end{bmatrix}H_{I}(x_{0},y_{0})\begin{bmatrix}
 x-x_0  \\ y-y_0 \end{bmatrix}
\nonumber
\end{align}
\noindent where $H_{I}(x_{0},y_{0}))$ is the Hessian matrix representing the second derivative values of $I$ at the specific point $(x_{0},y_{0})$. In order to have a better dividing path than a straight line connecting two contour points $(p, q)$, we start off from one end point $p$ and look for the next adjacent pixel with its local neighboring intensity change almost zero in one direction and drastically increased in its orthogonal direction. As a natural dividing curve passes along pixels with local intensity minimum, we can detect such adjacent pixels by computing two eigenvalues of Hessian matrix $H_{I}$ satisfying such properties: $0\approx \lambda_1 \ll \lambda_2$. The adjacent pixel with a near zero $\lambda_1$ and the maximum $\lambda_2$ is considered as the next pixel on the dividing curve.

To ensure that the final optimized connection path can merge to the other end point $q$ smoothly, we search forward in a targeting region with a constrained deviation angle (i.e. $\pm45^\circ$ in our experiment) from the vector connecting the current dividing point to $q$ at each step. Within such a sector-shaped searching space, we can force the resulting optimal connection path to remain in intensity ``valley'' with  large gradient variance and to converge smoothly to the other end point $q$.  This is demonstrated by the connection between point 5 and 9 in Fig.~\ref{fig:procedure}(\Romannum{5}) where our method is able to recover a dividing curve along a local minimum intensity route for overlapped nuclei segmentation.

\section{EXPERIMENTAL RESULTS}~\label{sec:result}
To validate the performance of the proposed algorithm, we tested our method with a dataset of 112 images ($1024 \times 1376$ pixels) from two tumor regions of seven brain tumor patients. As such images contain a large number of nuclei that are not overlapped, five image patches with overlapped nuclei clusters are randomly selected from each image. This results in 560 image patches for method evaluation. 

\begin{figure}
\centering
  \includegraphics[width=\linewidth,height=10.5cm]{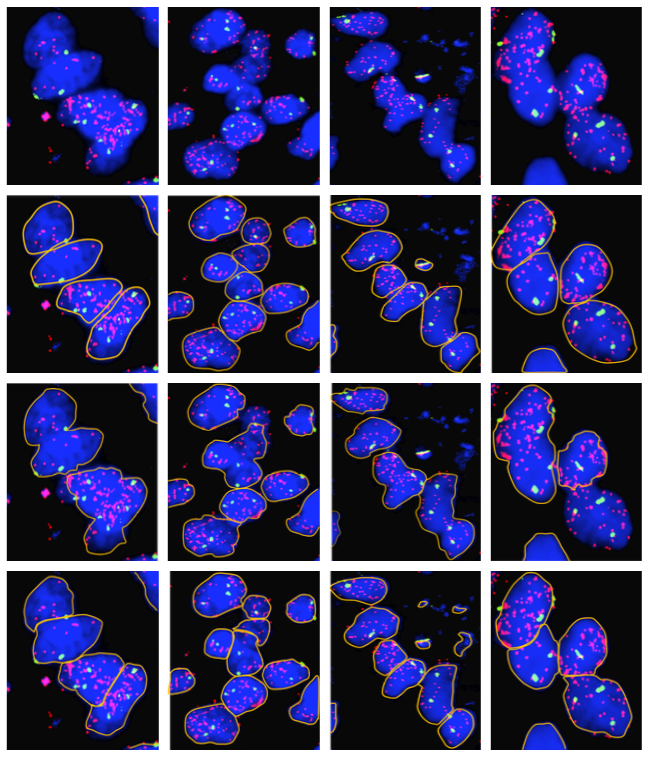}
  \caption{Comparisons of segmentation results over four representative overlapped nuclei regions are presented with original images, ground truth, results of marker-controlled watershed method, and outcomes of our proposed method from top to bottom rows, respectively.}
  \label{fig:res_compare} \inv\inv
\end{figure}

We visually inspect the resulting nuclei segmentation results and compare our method with the marker-controlled watershed method~\cite{watershed-based2017}. Representative results from four overlapped nuclei clusters in our dataset are presented in Fig.~\ref{fig:res_compare}. We present instances of clumped nuclei from original images in the first row, and ground-truth annotations on the second. Nuclei segmentation results from marker-controlled watershed segmentation and our proposed method are presented in the third and forth rows, respectively. It is noticed that our method is able to produce segmentation results similar to human annotations. In particular, those dividing curves recovered from local shape based intensity analysis can help improve segmentation results. By contrast, watershed method fails to identify and separate some overlapped nuclei clusters in the third row.

To quantitatively evaluate the performance of our proposed method, five metrics including Jaccard index, Precision, Recall, F1 score, and Hausdroff distance are used. The resulting quantitative evaluation results are presented in Table~\ref{tab:comparison}. 
For each patient, both marker-controlled watershed method and our proposed method are tested on 40 images of {\it Bulk Tumor} (BT) and another 40 of {\it Tumor Margin} (TM). There are four rows of results associated with each patient in Table~\ref{tab:comparison}. The top two lines present quantitative results of watershed based method  on BT and TM images, while the bottom two demonstrate results from our method on BT and TM images. Note that our method presents superior performance as measured by most metrics. Specifically, Jaccard index values from our proposed method are much higher than those of watered based method while Watershed method has better Precision as it tends to miss a large number of nuclei segregation events. Since our proposed method relies on the contour shape of the clustered nuclei,bit fails to work for the case where overlapped nuclei result in internal holes.
\begin{table}[h!]
\centering
\caption{Quantitative method evaluation and comparison with 560 image patches from BT and TM tissue regions: (mean,std).}
\label{tab:comparison}

\begin{tabular}{|c|c|c|c|c|c|}
\hline
\multicolumn{1}{|l|}{Patient} & Jaccard            & Precision          & Recall             & F1 score           & HD (pix.)                 \\ \hline
\multirow{4}{*}{1}            & 0.44,0.19          & \textbf{0.96,0.08}          & 0.46,0.21          & 0.58,0.22          & 34.3,19.1          \\ \cline{2-6} 
                              & 0.38,0.18          & \textbf{0.97,0.07}          & 0.40,0.20          & 0.53,0.20          & 34.6,18.7          \\ \cline{2-6} 
                              & \textbf{0.75,0.20} & 0.86,0.18 & \textbf{0.86,0.17} & \textbf{0.84,0.17} & \textbf{22.0,25.6} \\ \cline{2-6} 
                              & \textbf{0.81,0.14} & 0.93,0.13 & \textbf{0.87,0.11} & \textbf{0.89,0.11} & \textbf{15.0,19.6} \\ \hline
\multirow{4}{*}{2}            & 0.60,0.16          & \textbf{0.96,0.11}          & 0.62,0.16          & 0.73,0.16          & 20.2,15.5          \\ \cline{2-6} 
                              & 0.64,0.16          & \textbf{0.98,0.07}          & 0.66,0.16          & 0.77,0.13          & 24.3,17.9          \\ \cline{2-6} 
                              & \textbf{0.80,0.15} & 0.90,0.13 & \textbf{0.89,0.12} & \textbf{0.88,0.11} & \textbf{13.8,18.4} \\ \cline{2-6} 
                              & \textbf{0.83,0.15} & 0.91,0.12 & \textbf{0.91,0.12} & \textbf{0.90,0.11} & \textbf{13.6,20.1} \\ \hline
\multirow{4}{*}{3}            & 0.49,0.18          & \textbf{0.94,0.11}          & 0.52,0.21          & 0.63,0.19          & 32.2,23.5          \\ \cline{2-6} 
                              & 0.54,0.17          & \textbf{0.97,0.10}          & 0.55,0.17          & 0.68,0.15          & 31.7,20.3          \\ \cline{2-6} 
                              & \textbf{0.70,0.23} & 0.89,0.14 & \textbf{0.79,0.25} & \textbf{0.80,0.21} & \textbf{21.7,24.3} \\ \cline{2-6} 
                              & \textbf{0.77,0.17} & 0.89,0.11 & \textbf{0.85,0.16} & \textbf{0.86,0.13} & \textbf{13.8,18.2} \\ \hline
\multirow{4}{*}{4}            & 0.44,0.18          & \textbf{0.94,0.13}          & 0.46,0.20          & 0.59,0.20          & 33.6,22.0          \\ \cline{2-6} 
                              & 0.47,0.15          & \textbf{0.97,0.09}         & 0.48,0.16          & 0.62,0.16          & 34.5,17.9          \\ \cline{2-6} 
                              & \textbf{0.75,0.18} & 0.80,0.19 & \textbf{0.93,0.11} & \textbf{0.84,0.14} & \textbf{27.1,29.3} \\ \cline{2-6} 
                              & \textbf{0.80,0.19} & 0.87,0.15 & \textbf{0.92,0.16} & \textbf{0.88,0.15} & \textbf{22.0,29.0} \\ \hline
\multirow{4}{*}{5}            & 0.57,0.16          & \textbf{0.95,0.10 }         & 0.59,0.17          & 0.71,0.15          & 28.9,19.3          \\ \cline{2-6} 
                              & 0.49,0.16          & \textbf{0.97,0.10}          & 0.50,0.16          & 0.64,0.17          & 36.4,19.6          \\ \cline{2-6} 
                              & \textbf{0.80,0.17} & 0.85,0.16 & \textbf{0.93,0.13} & \textbf{0.88,0.14} & \textbf{21.4,30.8} \\ \cline{2-6} 
                              & \textbf{0.86,0.12} & 0.90,0.10 & \textbf{0.95,0.09} & \textbf{0.92,0.08} & \textbf{15.5,20.4} \\ \hline
\multirow{4}{*}{6}            & 0.51,0.19          & \textbf{0.93,0.13 }         & 0.55,0.21          & 0.65,0.19          & 26.6,19.5          \\ \cline{2-6} 
                              & 0.67,0.20          & \textbf{0.93,0.17 }         & 0.71,0.20          & 0.78,0.19          & 22.2,24.7          \\ \cline{2-6} 
                              & \textbf{0.71,0.21} & 0.84,0.18 & \textbf{0.84,0.20} & \textbf{0.81,0.17} & \textbf{23.8,25.4} \\ \cline{2-6} 
                              & \textbf{0.82,0.14} & 0.88,0.13 & \textbf{0.93,0.09} & \textbf{0.90,0.10} & \textbf{13.1,18.1} \\ \hline
\multirow{4}{*}{7}            & 0.58,0.20          & \textbf{0.95,0.11}          & 0.61,0.21          & 0.71,0.19          & 33.3,25.0          \\ \cline{2-6} 
                              & 0.44,0.20          & \textbf{0.96,0.08}          & 0.45,0.21          & 0.58,0.22          & 33.0,18.7          \\ \cline{2-6} 
                              & \textbf{0.78,0.18} & 0.86,0.14 & \textbf{0.91,0.16} & \textbf{0.86,0.14} & \textbf{19.2,30.3} \\ \cline{2-6} 
                              & \textbf{0.78,0.17} & 0.86,0.15 & \textbf{0.90,0.14} & \textbf{0.86,0.14} & \textbf{18.5,26.1} \\ \hline
\end{tabular}
\end{table}

\section{CONCLUSIONS}~\label{sec:conclusion}
This paper presents a novel segmentation method for clumped nuclei in fluorescence microsopy images. Our analysis first generates precise candidate point representations based on a high curvature point voting method, followed by detection of connecting point pairs based on spatial proximity, shape convexity, and curvature information via close point pair screening. An ellipse model is proposed to fit to each resulting area associated with each point pair candidate. We define a fitting quality as a function of  fitting angle, fitting area, fitting ellipse center shift, fitting perimeter change, and fitting area elongation. Only point pairs presenting good fitting quality are connected. Instead of separating overlapped nuclei by a straight line, we recover dividing curves by local shape based intensity analysis in a sector-shaped searching space. We validate our algorithm with 560 image patches from two classes of tumor regions associated with seven brain tumor patients. Both qualitative and quantitative validation results suggest that our algorithm is promising for dividing overlapped nuclei in fluorescence microscopy images widely used in various biomedical research. 

\addtolength{\textheight}{-12cm}   









\begin{thebibliography}{99}
\bibitem{Multifractal2017}Reljin N.,Slavkovic-Ilic M., Tapia C.,et al.,Multifractal-based nuclei segmentation in fish images. Biomedical microdevices, 19(3):2017

\bibitem{Delaunay2009}Wen, Q., Chang, H., and Parvin, B., A Delaunay triangulation approach for segmenting clumps of nuclei. IEEE International Symposium on Biomedical Imaging: From Nano to Macro, pp. 9-12, 2009.



\bibitem{bottleneck2011}Wang, H., Zhang, H. and Ray, N., September. Clump splitting via bottleneck detection. IEEE International Conference on Image Processing, pp. 61-64, 2011. 

\bibitem{Liao2016}Liao, M., Zhao, Y.Q., Li, X.H., Dai, P.S., Xu, X.W., Zhang, J.K. and Zou, B.J., Automatic segmentation for cell images based on bottleneck detection and ellipse fitting. Neurocomputing, 173, pp.615-622, 2016.





\bibitem{Fouad2016}Fouad, Shereen, et al. "Morphological Separation of Clustered Nuclei in Histological Images." International Conference Image Analysis and Recognition. Springer, Cham, 2016.

\bibitem{curvatureweighting2012}Zhang, C., Sun, C., Su, R. and Pham, T.D., Segmentation of clustered nuclei based on curvature weighting. Conference on Image and Vision Computing New Zealand, pp. 49-54, 2012.

\bibitem{Zhang2013}Zhang, C., C. Sun, and Tuan D. Pham. "Segmentation of clustered nuclei based on concave curve expansion." Journal of microscopy 251.1 (2013): 57-67.

\bibitem{Atta-Fosu}Atta-Fosu,Thomas,et al."3D Clumped Cell Segmentation Using Curvature Based Seeded Watershed."Journal of imaging 2.4(2016):31.
\bibitem{watershed-based2017}Bartell, L.R., Bonassar, L.J. and Cohen, I.,A watershed-based algorithm to segment and classify cells in fluorescence microscopy images. arXiv:1706.00815, 2017.

\end{thebibliography}
\end{document}